\documentclass{article}

\usepackage[preprint]{preprint}

\usepackage[utf8]{inputenc}
\usepackage[T1]{fontenc}
\usepackage{hyperref}
\usepackage{url}
\usepackage{booktabs}
\usepackage{amsfonts}
\usepackage{amsmath}
\usepackage{amssymb}
\usepackage{microtype}
\usepackage{graphicx}
\usepackage{array}
\usepackage{xcolor}
\usepackage{float}
\usepackage{listings}
\usepackage{algorithm}
\usepackage{algpseudocode}
\usepackage{fontawesome5}

\newcommand{\github}{\faGithub}
\newcommand{\pypi}{\faPython}
\newcommand{\website}{\faGlobe}
\newcommand{\ghlink}{https://github.com/MIT-LCP/croissant-baker}
\newcommand{\pypilink}{https://pypi.org/project/croissant-baker/}
\newcommand{\websitelink}{https://lcp.mit.edu/croissant-baker/}

\definecolor{jsonkw}{rgb}{0.13,0.13,0.60}
\definecolor{jsonval}{rgb}{0.13,0.50,0.13}
\lstdefinelanguage{json}{
  basicstyle   = \ttfamily\scriptsize,
  morestring   = [b]",
  stringstyle  = \color{jsonval},
  literate     = *{:}{{{\color{jsonkw}:}}}{1},
  breaklines   = true,
  columns      = flexible,
  keepspaces   = true,
  showstringspaces = false,
  frame        = single,
  framerule    = 0.4pt
}

\title{Croissant Baker: Metadata Generation for Discoverable, Governable, and Reusable ML Datasets}

\author{%
\parbox{\linewidth}{\centering
Rafi Al Attrach$^{1,2,\ast}$, Rajna Fani$^{1,2}$, Sebastian Lobentanzer$^{3}$, Joan Giner-Miguelez$^{4}$, Debanshu Das$^{5}$, Varuni H.\,K.$^{6}$, Nobin Sarwar$^{7}$, Rajat Ghosh$^{8}$, Anwai Archit$^{9}$, Surbhi Motghare$^{10}$, Christina Conrad Parry$^{11}$, Luis Oala$^{12}$, Lara Grosso$^{13}$, Joaquin Vanschoren$^{14}$, Steffen Vogler$^{15}$, Sujata Goswami$^{16}$, Eric S.\ Rosenthal$^{17}$, Marzyeh Ghassemi$^{2}$, Matthew McDermott$^{18}$, Tom Pollard$^{2}$
\\[0.8ex]
{\normalfont\mdseries\small
$^{1}$Technical University of Munich, $^{2}$Massachusetts Institute of Technology, $^{3}$Helmholtz Munich, $^{4}$Barcelona Supercomputing Center, $^{5}$Google, $^{6}$Couchbase, $^{7}$University of Maryland, Baltimore County, $^{8}$Nutanix, $^{9}$Georg-August-University G\"ottingen, $^{10}$Salesforce, $^{11}$Sage Bionetworks, $^{12}$Dotphoton, $^{13}$Harvard University, $^{14}$Eindhoven University of Technology, $^{15}$Bayer AG, $^{16}$Independent Researcher, $^{17}$Massachusetts General Hospital, $^{18}$Columbia University
\\[0.4ex]
$^{\ast}$Correspondence: \texttt{rafiaa@mit.edu}
}}}

\begin{document}

\maketitle

\begin{center}
\small
\begin{tabular}{@{}r@{\hspace{0.6em}}l@{}}
\github  & \url{\ghlink} \\
\pypi    & \url{\pypilink} \\
\website & \url{\websitelink} \\
\end{tabular}
\end{center}

\begin{abstract}
Croissant has emerged as the metadata standard for machine learning datasets, providing a structured, JSON-LD--based format that makes dataset discovery, automated ingestion, and reproducible analysis machine-checkable across ML platforms. Adoption has accelerated, and NeurIPS now requires Croissant metadata in every submission to its dataset tracks. Yet in practice Croissant generation usually starts with uploading data to a public platform, a path infeasible for governed and large local repositories that hold much of the high-value data ML increasingly relies on. We release Croissant Baker, a local-first, open-source command-line tool that generates validated Croissant metadata directly from a dataset directory through a modular handler registry. We evaluate Croissant Baker on over 140 datasets, scaling to MIMIC-IV at 886 million rows and 374 Parquet files. On held-out comparisons against producer-authored or standards-derived ground truth, Croissant Baker reaches 97--100\% agreement across multiple domains.
\end{abstract}

\section{Introduction}

Croissant~\citep{wilkinson2016fair,akhtar2024croissant} has emerged as the metadata standard for machine learning datasets: a JSON-LD--based format that makes datasets directly loadable into ML frameworks such as TensorFlow Datasets, PyTorch, and JAX, indexable by schema.org-aware dataset search engines, and verifiable as packaging contracts before ingestion. Adoption now spans hundreds of thousands of datasets across Hugging Face, Kaggle, OpenML, and Google Dataset Search~\citep{mlcommons2025momentum}, and Croissant has become a submission requirement for new datasets at venues such as NeurIPS~\citep{neurips2025hosting}. Beyond ML tooling, Croissant also bridges to scientific domains with decades of ontology tradition~\citep{bioschemas}, and complements regulatory frameworks such as the European Health Data Space~\citep{eu2022ehds} and forthcoming U.S.\ HTI-5 rules that mandate structured, machine-readable representations of high-stakes data. Croissant has, in short, become the connective tissue between ML, scientific, and regulatory data ecosystems.

Yet the current authoring pipeline silently assumes that data upload is required. Hugging Face, Kaggle, and OpenML all generate metadata only after a dataset has been transmitted to a public platform, a path that is infeasible for the data ecosystems where ML increasingly matters: clinical data governed by HIPAA and DUAs~\citep{physionet_dua}, government data subject to procurement and security boundaries, and enterprise data locked by NDAs and IP concerns. What these settings need is local-first generation, where metadata is produced from a dataset directory in place, without transmission to a third-party service; no current authoring path provides this. Even when upload is permitted, platform-side generation cannot fully close the gap, because turning files into valid Croissant requires a \emph{recovery layer} between raw bytes and dataset structure that no generic file walk reconstructs. WFDB records~\citep{xie2023wfdb} couple \texttt{.hea} headers with one or more \texttt{.dat} signal files; FHIR~\citep{hl7fhir} requires content-aware dispatch between Bundle and NDJSON serializations; Parquet tables may be partitioned across directories; DICOM~\citep{dicom_ps36} and NIfTI encode acquisition metadata in headers that must be parsed without materializing pixel or voxel payloads; multi-band scientific TIFFs (e.g.\ 12-band Sentinel-2) carry band structure that ordinary image abstractions collapse. These operations (summarized visually in Figure~\ref{fig:format_complexity}) encode domain-specific knowledge that platform generators do not, leading them to misclassify datetimes as dates, integer IDs as text, or emit empty schemas for waveform and multi-band inputs (Appendix~\ref{app:hf_kaggle}).

Recent agent-assisted approaches address a different part of the problem. The \texttt{eclair} Model Context Protocol (MCP) server~\citep{mlcommons2025eclair} and \texttt{mlcbakery}~\citep{jetty2025mlcbakery} support iterative Croissant creation through MCP client interfaces focused on semantic annotation, which is well-suited to natural-language fields such as description, citation, and intended use, but cannot on its own recover format-specific structure from local files and does not by default operate entirely offline. Direct LLM-only generation likewise faces three structural obstacles independent of accuracy on text: outputs are not deterministically derivable from source files (a reproducibility blocker for governed data), context windows do not accommodate institutional-scale repositories such as a multi-gigabyte Parquet store, and frontier models do not natively parse binary headers in DICOM, NIfTI, multi-band TIFF, or WFDB without dispatching to format-specific tools. The gap is therefore twofold: governance excludes the data, and neither platform-side generation nor agent-assisted annotation alone recovers the structure once you have it locally.

We present Croissant Baker, an open-source tool that closes this gap with two architectural commitments. First, a clean separation between deterministic structural inference (file-derived, byte-traceable, reproducible by construction) and semantic enrichment (CLI- or agent-supplied under explicit human review). Every value in the output Croissant document is therefore auditable to either source-file bytes or an explicit input, and the structural core composes with agent-based authoring rather than competing with it. Second, a typed handler protocol that addresses the long tail of scientific and domain-specific formats: new formats register once with the dispatch table without modifying the inference core, exercised by regression tests on fixture datasets. Built-in handlers span tabular, columnar, JSON, waveform, image, and biomedical formats including WFDB, DICOM, NIfTI, FHIR, OMOP, and MEDS. Our contributions are summarized as follows:
\begingroup
\setlength{\topsep}{0pt}
\setlength{\partopsep}{0pt}
\begin{itemize}
\setlength{\itemsep}{2pt}
\setlength{\parskip}{0pt}
\item \textbf{Format-aware structural recovery.} We identify recovering valid dataset structure from heterogeneous scientific and domain-specific file formats as the central obstacle to scalable Croissant authoring, and introduce a structural/semantic separation that keeps structural inference deterministic and auditable while remaining compatible with agent-assisted enrichment (\S\ref{sec:format_depth}, Appendix~\ref{app:agent_extension}).
\item \textbf{Open-source local-first tool.} We release Croissant Baker as an open-source, local-first implementation, evaluated on 140+ datasets and scaling to MIMIC-IV at 886M rows across 374 Parquet files (\S\ref{sec:results}).
\item \textbf{Held-out standards-grounded validation.} We provide held-out validation against producer-authored and standards-grounded ground truth: 97.9\% semantic type agreement on a deterministic seeded draw of 25 datasets across the 11 OpenReview primary-area buckets of the NeurIPS 2025 D\&B track, 97.4\% on 55 Open Targets datasets, 97.8\% on a SMART Health IT FHIR release resolved against US Core STU7 and HL7 R4, and 100\% strict tag-ID agreement against the DICOM PS3.6 dictionary across six vendor modules (\S\ref{sec:heldout_summary}).
\end{itemize}
\endgroup

Our evaluation emphasizes biomedical and adjacent scientific datasets because they combine the two pressures that make Croissant authoring hard in practice: strict governance constraints that rule out upload, and substantial format diversity that rules out one-size-fits-all parsing. Once a dataset carries valid Croissant, downstream consumers can load it without dataset-specific ingestion code, compare schemas across pipelines for federated or multi-site analysis, and use the document as a packaging contract that detects file or schema drift before training. By bringing governed and long-tail data into ML metadata standards, Croissant Baker expands the set of datasets that can meet emerging publication and review requirements while unlocking these downstream uses for the data ecosystems where ML increasingly matters.

\section{Methods}

\subsection{Croissant Specification}

Croissant is a JSON-LD--based metadata specification built upon the Schema.org vocabulary~\citep{mlcommons_spec}. It represents datasets through four primary components: (1)~dataset-level metadata (name, description, license, creators); (2)~file distributions enumerating file-level FileObject resources with location and format; (3)~one or more RecordSet objects defining the logical structure of structured data including field names and data types; and (4)~ML semantics encoding train/test/validation splits and label assignments. Data types map to Schema.org primitives (e.g.\ \texttt{sc:Integer}, \texttt{sc:Float}, \texttt{sc:Text}, \texttt{sc:Date}), enabling downstream tools to anticipate data characteristics prior to ingestion. 

Croissant~1.1 has been released~\citep{mlcommons2026croissant11}, extending the specification with additional semantic fields and dataset-linking capabilities. Croissant Baker targets Croissant~1.1 and passes through native Responsible AI (RAI) metadata when those fields are provided by the user.

\subsection{Why Format-Aware Handlers Matter}
\label{sec:format_depth}

\begin{figure}[t]
    \centering
    \includegraphics[width=\linewidth]{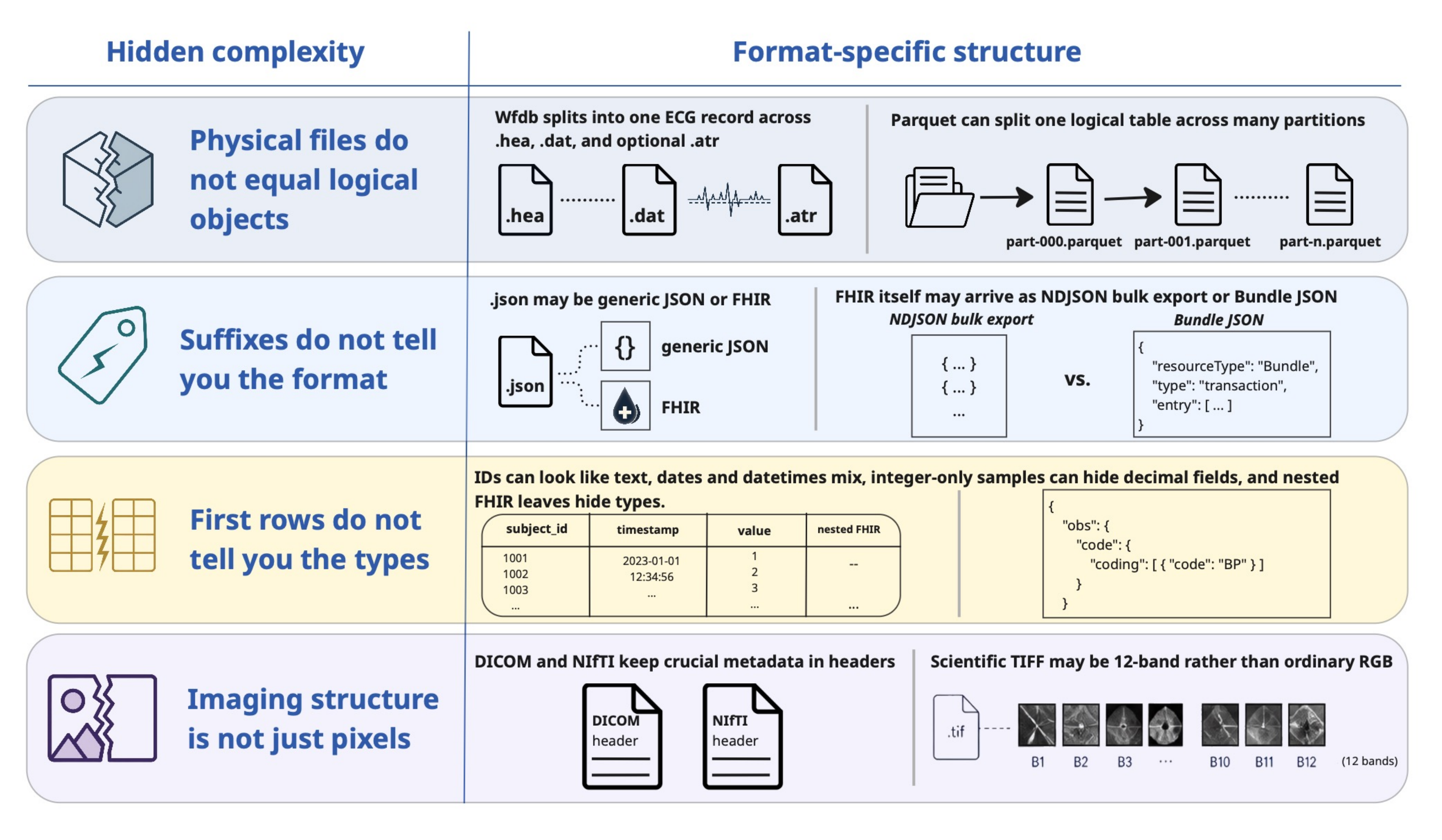}
    \caption{\textbf{Format-specific structure that a generic file walk does not recover.} The left column shows simplifying assumptions that hold for ordinary file inspection but fail on scientific datasets. The right column shows the structure Croissant Baker must recover before it can emit valid Croissant metadata: WFDB records span coupled \texttt{.hea}/\texttt{.dat}/\texttt{.atr} files, Parquet tables may be partitioned across directories, FHIR requires content-aware dispatch between Bundle and NDJSON serializations, types can change after deeper sampling or nested expansion, and DICOM, NIfTI, and multi-band TIFF encode crucial metadata in headers or non-RGB band layouts. These recovery steps explain the gap between Croissant Baker and platform auto-generation.}
    \label{fig:format_complexity}
\end{figure}

The technical challenge in Croissant Baker sits in the recovery layer between raw files and valid dataset structure. A directory walk enumerates paths, suffixes, and byte sizes. It does not tell us which files belong to one logical record, whether a \texttt{.json} file is generic JSON or FHIR, whether early rows expose the final schema, or how to recover imaging metadata without loading dense payloads (See Figure~\ref{fig:format_complexity}). The Waveform Database Format (WFDB) needs \texttt{.hea} headers before its waveform files admit a schema; FHIR needs content sniffing on \texttt{resourceType} followed by nested-structure expansion; Parquet needs schema-only inspection and partition regrouping into one logical table; DICOM and NIfTI need header-only reads so geometry and acquisition metadata are recovered without materializing pixel or voxel payloads; and multi-band TIFF needs a dedicated decoding path to preserve scientific band structure. These operations explain the common failure modes of platform-generated Croissant we observe in Appendix~\ref{app:hf_kaggle} (datetimes misclassified as dates, integer IDs as text, empty schemas for WFDB and multi-band TIFF) and motivate the typed handler protocol: new handlers register with the dispatch table without changes to the inference core, with regression tests on fixture datasets guarding the contract.

\subsection{Implementation}

Croissant Baker is distributed as an open-source Python library on PyPI as \texttt{croissant-baker}. The CLI accepts a dataset directory path with optional metadata overrides for core fields (name, description, creator, citation, license) and Croissant~1.1 / Schema.org fields (publisher, \texttt{sameAs}, temporal coverage, usage information, alternate names, version, native RAI attributes). Users may additionally supply field mappings to attach \texttt{equivalentProperty} links or extra data-type URIs after structural inference. A representative end-to-end invocation and its JSON-LD output are shown in Appendix~\ref{app:cli_example} and Appendix~\ref{app:jsonld}.

The Croissant Baker implementation handles repositories at large scale through a four-stage sequential pipeline; total generation time is determined by the combined cost of these stages.

\begin{enumerate}
  \item \textbf{File Discovery.} Recursive traversal discovers all files while preserving their relative paths. SHA-256 checksums and file sizes are computed over raw bytes on disk, including compressed files. Hashing compressed bytes preserves the exact file state as downloaded or archived and enables independent verification with standard tools.
  \item \textbf{Handler Dispatch.} Files are matched to format-specific handlers via a registry pattern.
  \item \textbf{Metadata Extraction.} Handlers analyze file contents to extract structural information: column names, inferred types, and row counts for tabular files; logical record groupings for multi-file formats; dimensions, band count, and format for images.
  \item \textbf{Croissant Generation.} Extracted metadata are assembled into Croissant objects (Metadata, FileObject, RecordSet, Field), serialized to JSON-LD, and validated using the \texttt{mlcroissant} library.
\end{enumerate}

\begin{figure}[t]
    \centering
    \includegraphics[width=\linewidth]{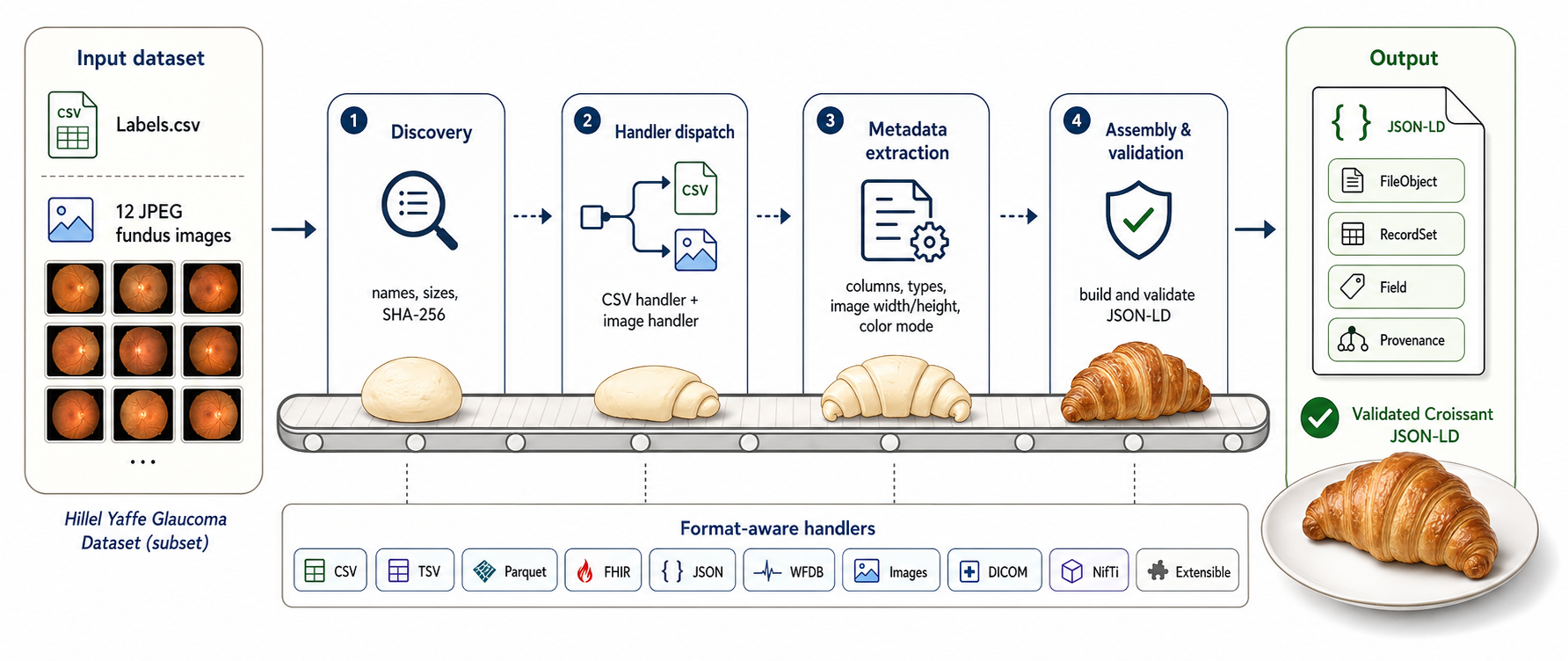}
    \caption{\textbf{Croissant Baker technical architecture.} Files are discovered locally, hashed, and dispatched through an ordered handler registry to the built-in CSV/TSV, Parquet, FHIR, JSON/JSONL, WFDB, image, DICOM, and NIfTI handlers. Handler outputs are assembled into Croissant \texttt{FileObject}, \texttt{FileSet}, \texttt{RecordSet}, and \texttt{Field} resources, then validated with the \texttt{mlcroissant} library. New formats are supported by implementing the handler protocol and registering once with the registry.}
  \label{fig:architecture}
\end{figure}

\subsubsection{Format Handlers}

Croissant Baker ships an extensible set of built-in handlers; full per-handler details are in Appendix~\ref{app:format_handlers}.
\textbf{CSV/TSV} (with \texttt{.gz}, \texttt{.bz2}, \texttt{.xz} variants) infers column types via Apache Arrow with temporal-field pattern recognition; covers OMOP datasets distributed as compressed CSVs.
\textbf{Parquet} reads Arrow schema metadata without loading full datasets and covers MEDS~\citep{mcdermott2025meds}.
\textbf{FHIR}~\citep{hl7fhir} supports both NDJSON bulk export (one resource per line) and JSON documents (Bundle or single resource), with content sniffing on \texttt{resourceType}.
\textbf{JSON/JSONL} handles generic non-FHIR JSON arrays, single objects, and JSON Lines (with gzip variants), inferring schema from inspected records.
\textbf{WFDB}~\citep{xie2023wfdb} parses \texttt{.hea} headers to recover sampling frequency, lead names, and record structure, then groups \texttt{.hea}/\texttt{.dat}/\texttt{.atr} files into a RecordSet describing the logical signal; this file-level coupling is structure that platform tools fail to encode (Table~\ref{tab:hf_comparison}).
\textbf{Image} (JPEG, PNG, TIFF, GIF, BMP, WebP) extracts dimensions, band count, format, and SHA-256; multi-band scientific TIFFs (e.g.\ 12-band Sentinel-2) route through \texttt{tifffile} for multimodal datasets.
\textbf{DICOM} reads \texttt{.dcm}/\texttt{.dicom} files with \texttt{pydicom} in header-only mode (\texttt{stop\_before\_pixels}), extracting modality, geometry, frame count, encoding, and study/series identifiers without materializing pixel arrays.
\textbf{NIfTI} parses \texttt{.nii}/\texttt{.nii.gz} headers via \texttt{nibabel}, recovering spatial dimensions, voxel spacing, NIfTI version, and 4D repetition time.

\subsection{Evaluation Methodology}
\label{sec:eval_methodology}

We organize the evaluation into seven splits: two local splits that establish coverage and scale, and five held-out splits that test agreement with reference metadata, generalization to independent corpora, and cross-domain coverage of the NeurIPS 2025 Datasets and Benchmarks track. Appendix Table~\ref{tab:eval_splits} lists each split with its corpus, ground truth, and supported claim.


The two local splits use nine datasets that cover compressed clinical CSV, Parquet event streams, WFDB waveform records, multimodal image + tabular collections, and full-scale institutional repositories. Appendix~\ref{app:local_datasets} details each dataset and its role. Seven smaller datasets are used during development to exercise the CSV/TSV, Parquet, WFDB, and image handlers end-to-end. Two full-scale repositories, MIMIC-IV CSV and MIMIC-IV MEDS, are reserved for runtime and repository-scale validation, as they share schemas with their smaller counterparts.

The five held-out splits use corpora distinct from those exercised during development. The NeurIPS 2025 cross-domain split applies a deterministic seeded draw across the 11 OpenReview primary-area buckets summarized in Figure~\ref{fig:dandb_landscape}, evaluating Croissant Baker against the Croissant snapshots submitted alongside each paper at review time. Open Targets provides 55 Parquet datasets with producer-authored Croissant metadata, enabling evaluation of RecordSet recovery, field-name recovery, and type agreement against ground truth. SMART Health IT FHIR NDJSON and JSON Bundle files evaluate the FHIR handler on an independent source, standard resource naming, and a second ingestion path. For medical imaging, the \texttt{dcm\_validate}~\citep{rorden2025dcmvalidate} corpus pairs vendor DICOM acquisitions with NIfTI conversions and BIDS sidecars. A separate set of 51 publicly available BIDS datasets from OpenNeuro~\citep{openneuro2022} evaluates the NIfTI handler at scale across heterogeneous independent datasets.

For the local splits, we assess execution success, \texttt{mlcroissant} validity, structural fidelity of file/RecordSet counts, and downstream utility. For the held-out splits, let $F_g$ and $F_r$ denote the field sets of a generated and a reference Croissant document, and $M = F_g \cap F_r$ the matched fields after namespace normalization. Field-name recovery, strict type agreement, and semantic type agreement are
\[
  R_{\mathrm{field}} = \frac{|M|}{|F_r|}, \qquad
  T_{\mathrm{strict}} = \frac{1}{|M|}\sum_{f \in M}\mathbf{1}\!\left[\tau_g(f) = \tau_r(f)\right], \qquad
  T_{\mathrm{sem}} = \frac{1}{|M|}\sum_{f \in M}\mathbf{1}\!\left[\tau_g(f) \sim \tau_r(f)\right],
\]
where $\tau_g, \tau_r$ map fields to Croissant types and $\sim$ holds within the same numeric family. RecordSet-name recovery is defined analogously over RecordSet names. Timing benchmarks are executed on a MacBook Pro with an Apple M1~Max processor (10 cores) and 32\,GB RAM (full specification in Appendix~\ref{app:hardware}).

\section{Results}
\label{sec:results}

\subsection{Development and Scalability Results}

Croissant Baker generates valid metadata for all nine development and scalability datasets, spanning tabular, columnar, waveform, and image modalities. Across these datasets, 768 files are represented as FileObject entries with 599 logical RecordSet objects, and all outputs pass \texttt{mlcroissant} validation without modification. Generation time ranges from 0.74\,s on Glaucoma Fundus (13 files) to 32.2\,s on MIMIC-IV MEDS full (366 Parquet files, 3.67\,GB); the two full-scale runs (MIMIC-IV full at 9.92\,GB in 13.3\,s and MIMIC-IV MEDS full at 3.67\,GB in 32.2\,s) demonstrate scalability to institutional-scale repositories. Per-dataset counts and timings appear in Appendix Table~\ref{tab:results}.

\subsection{Structural Fidelity, Cross-Modal Versatility, and Comparison}

Structural fidelity is confirmed by matching file, table, and schema-field counts between source directories and generated Croissant. For WFDB, the 213 component files group into 71 logical records, with sampling frequency (360\,Hz) and lead names (MLII, V5) correctly extracted; for the 12-band Sentinel-2 dataset, 10 TIFF files and a 1017-column CSV are processed in one workflow, producing per-band image metadata and typed tabular RecordSets. No dataset-specific configuration is required across either local split.

Table~\ref{tab:comparison} contrasts Croissant Baker with existing approaches: manual authoring is high-effort (15--30\,min per dataset~\citep{akhtar2024croissant}), platform generation requires upload that is often infeasible under DUAs~\citep{huggingface_dpa}, and Croissant Baker provides automated local generation suitable for batch processing. On seven open-access datasets uploaded to Hugging Face and Kaggle for direct comparison (per-dataset results in Appendix~\ref{app:hf_kaggle}), Baker produces more granular file-level metadata (per-file SHA-256 + byte sizes vs.\ placeholder links or archive-level checksums), more precise inferred types (e.g., \texttt{cr:Int64}, \texttt{sc:DateTime}) for multi-band and multimodal datasets where the HF/Kaggle outputs are empty or coarse, and full provenance (license, citation, datePublished); MIMIC-IV and MIMIC-IV MEDS at full scale are excluded due to DUA restrictions.

\begin{table}[t]
\caption{Comparison of Croissant metadata generation approaches.}
\label{tab:comparison}
\centering
\small
\begin{tabular}{lcccc}
\toprule
\textbf{Capability} & \textbf{Manual} & \textbf{HF/Kaggle} & \textbf{Croissant Editor} & \textbf{Croissant Baker} \\
\midrule
DUA compliant        & Yes & No       & No       & Yes \\
Batch processing     & No  & Yes      & No       & Yes \\
Works offline        & Yes & No       & No       & Yes \\
Auto type inference  & No  & Yes      & Yes      & Yes \\
Time required        & High & Low (after upload) & Moderate & Low ($<$33\,s) \\
Code required        & Yes & No       & No       & No \\
User interface       & Text editor & Web UI & Web UI & CLI \\
API integration      & No  & Limited  & No       & Yes \\
\bottomrule
\end{tabular}
\end{table}

\paragraph{Downstream utility.} The \texttt{mlcroissant} Python API loads each generated document and iterates over RecordSets without dataset-specific ingestion code, yielding typed dictionaries for tabular data and logical ECG records with correct \texttt{.hea}/\texttt{.dat} file associations for WFDB. Cross-site schema comparison reduces to a single \texttt{mlcroissant} load and a set diff: an OMOP-to-MEDS or two-site MIMIC-IV consistency check that previously required custom parsers per institution becomes a one-liner over the Croissant documents. The same documents serve as packaging contracts: controlled perturbations (removed file, renamed waveform component, changed column name) are detected by \texttt{mlcroissant} validation before downstream analysis, and metadata-only export supports discoverability and submission compliance for controlled-access data without moving patient-level content. Appendix~\ref{app:downstream} provides full details.

\subsection{Held-Out Evaluation Summary}
\label{sec:heldout_summary}

Table~\ref{tab:heldout_summary} summarizes the five held-out evaluations, which collectively test cross-domain coverage of the NeurIPS 2025 D\&B track, agreement with producer-authored Croissant metadata, generalization to an independent FHIR source, cross-vendor imaging generalization, and NIfTI/BIDS coverage at scale.

\begin{table}[t]
\caption{Summary of held-out evaluation results.}
\label{tab:heldout_summary}
\centering
\scriptsize
\setlength{\tabcolsep}{4pt}
\begin{tabular}{@{}>{\raggedright\arraybackslash}p{0.17\linewidth}
                >{\raggedright\arraybackslash}p{0.26\linewidth}
                >{\raggedright\arraybackslash}p{0.23\linewidth}
                >{\raggedright\arraybackslash}p{0.23\linewidth}@{}}
\toprule
\textbf{Split} & \textbf{Corpus} & \textbf{Key result} & \textbf{Claim supported} \\
\midrule
NeurIPS 2025 cross-domain & 25 datasets across 11 OpenReview primary-area buckets, 3 seeds & 24/25 datasets produce valid Croissant; 97.9\% semantic type agreement on 235 fields against submission-time Croissant snapshots & Cross-domain coverage of the NeurIPS 2025 D\&B track \\
Open Targets & 55 producer-authored Parquet Croissant datasets & 55/55 RecordSet names, 819/819 field names, 798/819 semantic types (97.4\%) recovered & Agreement with external, producer-authored Croissant metadata \\
FHIR & SMART Health IT NDJSON bulk export and JSON Bundle files & 397/406 strict type matches (97.8\%) on the NDJSON release; all outputs valid for both ingestion paths & FHIR generalization on an independent source and a second ingestion path \\
Paired DICOM/NIfTI & \texttt{dcm\_validate}, 6 vendor modules & 6{,}678 DICOM files and 75 NIfTI volumes processed; 48/48 emitted DICOM tag identifiers resolved to PS3.6 keywords & Cross-vendor DICOM and NIfTI handler generalization \\
OpenNeuro & 51 publicly available BIDS datasets & 51/51 datasets baked without failure; all outputs passed Croissant~1.1 validation & NIfTI/BIDS coverage across heterogeneous independent datasets \\
\bottomrule
\end{tabular}
\end{table}

\subsection{NeurIPS 2025 Cross-Domain Evaluation}
\label{sec:neurips_2025_cross_domain}

\begin{figure}[!ht]
  \centering
  \includegraphics[width=\linewidth]{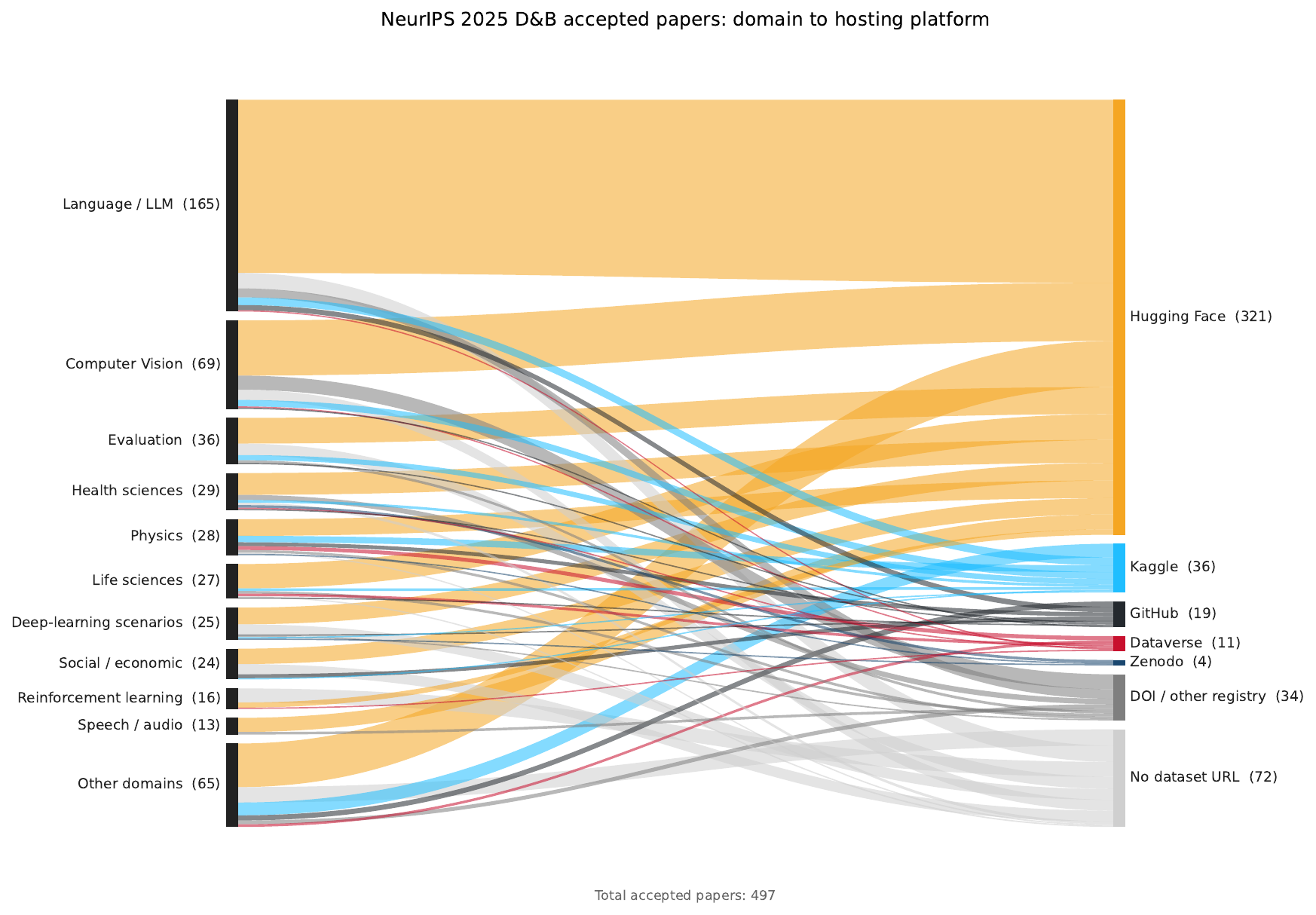}
  \caption{\textbf{NeurIPS 2025 Datasets and Benchmarks track at a glance.} Domain (left) to hosting platform (right) for the 497 accepted papers, taken from the OpenReview \texttt{primary\_area} and \texttt{dataset\_URL} fields. Hugging Face hosts approximately two thirds of the public datasets, while the remaining datasets are distributed across Kaggle, GitHub, Dataverse, Zenodo, and DOI registries. Notably, 14.5\% of accepted papers do not include a public dataset URL, often corresponding to method, framework, or benchmark-protocol papers. Appendix~\ref{app:dandb_composition} provides the full numerical breakdown.}
  \label{fig:dandb_landscape}
\end{figure}

To evaluate Croissant Baker on the actual distribution of datasets in the NeurIPS Datasets and Benchmarks track, we run a deterministic seeded draw against the OpenReview snapshot in Figure~\ref{fig:dandb_landscape}: 497 accepted papers grouped into the 11 \texttt{primary\_area} buckets that span all major scientific and applied-ML domains represented in the track. For reproducibility and to widen bucket coverage, we draw three independent seeds and restrict eligibility to publicly retrievable Hugging Face datasets, the track's most prevalent host with about two thirds of the public datasets in Figure~\ref{fig:dandb_landscape}. For each (seed, bucket), we deterministically shuffle the bucket's eligible candidates and select the first dataset that meets the size band; institutional-scale repositories are covered separately in the scalability split. The protocol resolves to 25 unique datasets across 33 picks. Appendix~\ref{app:neurips_2025_protocol} documents the eligibility criteria, the size band, and the seed values; the candidate pool and the shuffle are released with the source code.

For each pick, Croissant Baker generates metadata directly from the dataset directory and the result is compared against the Croissant snapshot bundled with the paper at submission time. Baker produces valid Croissant for 24 of 25 datasets; the one failure (UniHG, shipped as Apache Arrow rather than the more common Parquet) falls outside the current handler set and exits cleanly with ``no supported files found'' rather than emitting incorrect metadata, a natural extension target for the modular registry of Section~\ref{sec:format_depth} where new handlers register without changes to the structural inference core. Across the 235 fields shared between Baker output and the producer Croissants, semantic type agreement reaches 97.9\%, comparable to the 97.4\% reported on Open Targets (Section~\ref{sec:external_validation}) and the 97.8\% reported on FHIR (Section~\ref{sec:fhir_ood}).

The 11 buckets jointly cover language, vision, evaluation, health, physics, life sciences, deep-learning scenarios, social and economic aspects, reinforcement learning, speech and audio, and a residual ``other'' category. Agreement at this level across the full domain breadth of the track, not only the modality-specific subsets covered by the remaining held-out splits, supports the claim that the recovery layer described in Section~\ref{sec:format_depth} generalizes across the format heterogeneity the track contains.

Per-split details for Open Targets, FHIR (NDJSON and JSON Bundle), DICOM/NIfTI on the \texttt{dcm\_validate} corpus, and OpenNeuro at scale are reported in Appendix~\ref{app:heldout_eval_details}. Brief framing: the 21 Open Targets type disagreements (2.6\%) all reduce to producer Croissants assigning \texttt{sc:Float} to integer-valued columns where Baker infers the more precise integer from the Parquet schema. The 9 FHIR disagreements (2.2\%) reduce to eight URI-shaped fields where the specification declares \texttt{System.String} (\texttt{sc:Text}) and Baker infers the more specific \texttt{sc:URL}, plus one decimal field with only integer values in the 10-patient sample. The DICOM agreement against PS3.6 keywords is 48/48 strict; the OpenNeuro pass rate is 51/51 datasets validated without failure.

\section{Discussion}

Many existing Croissant generation workflows assume public hosting, browser-based interaction, or data upload. Controlled-access datasets often rule out those options. Upload-based tools also depend on network and compute bandwidth, which becomes a practical bottleneck for large repositories. Croissant Baker instead performs metadata generation entirely within the local compute environment.

\textbf{Broader applications and reproducibility.}
Although the most demanding cases in this paper come from biomedical data, the same local-first workflow extends to research groups, companies, and public data stewards that require Croissant metadata for datasets they cannot or do not want to upload during authoring. Practical examples include local validation of an external model on an institution's own cohort, as well as exchanging Croissant artifacts between sites running OMOP or MEDS pipelines to verify schema compatibility prior to federated analysis, without sharing patient-level data~\citep{wang2020mimicextract,tang2020fiddle}.

\textbf{Failure modes and limitations.}
Type inference relies on heuristics applied to sampled rows and may misclassify ambiguously encoded fields; manual review remains advisable for columns with irregular formatting. The Baker output also gives primary authors a fast feedback loop for resolving such ambiguities at the source: re-running the tool after a fix updates the inferred schema deterministically, so eliminating ambiguities in the raw files is the most sustainable path to clean Croissant metadata. Datasets in unsupported formats are skipped, and binary files without structured tabular data produce \texttt{FileObject} entries without RecordSets. Memory can become a constraint for very large files that require full loading. Inter-table relationships, such as foreign keys, are not inferred automatically; for relational schemas like OMOP, where concept tables and event tables are linked by \texttt{concept\_id}, the generated Croissant document captures per-table structure but not cross-table joins. Relationship inference is a natural extension target, and the agentic enrichment path described in Appendix~\ref{app:agent_extension} provides one route for proposing such links from schema documentation without modifying the deterministic structural core. Croissant describes structural metadata but not data-collection methodology or cohort definition; complementary frameworks, including Datasheets for Datasets~\citep{gebru2021datasheets} and Data Cards~\citep{pushkarna2022datacards}, remain necessary to document these aspects.

\textbf{Broader impacts.} Incorrectly inferred metadata could propagate to downstream ingestion, schema matching, or evaluation workflows, and richer metadata may increase visibility of sensitive datasets if combined with weak governance. For these reasons, we encourage human review of Croissant Baker outputs and recommend caution when using it in combination with agent-assisted semantic enrichment.

\textbf{Future work.}
Remaining format priorities include BAM, VCF, and other scientific containers. The FHIR handler currently supports NDJSON and JSON Bundle formats; future extensions will broaden coverage to additional dialect variants and enable cross-resource relationship inference. Another direction is dual compatibility between Croissant and Bioschemas profiles~\citep{bioschemas}; the BioCroissant working group~\citep{biocroissant} suggests modest changes to core Croissant fields can accommodate life-science ontology constraints. Finally, broader cross-domain evaluation on producer-authored or platform-authored Croissant metadata outside biomedical data would further strengthen the benchmark matrix.

\textbf{Agent-assisted enrichment.}
Semantic fields that Baker requires as CLI input (description, citation, license, creator) cannot be inferred safely from local files alone. Retrieval-augmented LLMs have shown practical utility for such fields~\citep{alyafeai2025mole, tinn2025premeta}, and an MCP-mediated agent~\citep{anthropic2024mcp, mlcommons2025eclair} can propose values from repository landing pages, PubMed, or DataCite without altering the deterministic structural pipeline (Appendix~\ref{app:agent_extension}). The two regimes are complementary: structural inference is reproducible by construction and free of per-invocation token cost; agentic enrichment is well-suited to natural-language fields. Croissant Baker can also be exposed as a local MCP server so agents access dataset directories without leaving the secure environment. LLM-generated text remains prone to factual error~\citep{huang2025hallucination}, so human review is a prerequisite for any proposed values.

\vspace*{-0.5ex}
\section{Conclusion}

Governed datasets are frequently excluded from ML metadata workflows that assume public hosting or third-party upload. Croissant Baker closes this gap with local-first, automated generation of specification-compliant Croissant metadata. The tool produces valid metadata across nine heterogeneous local datasets and reaches 97.9\% semantic type agreement on a deterministic seeded draw of 25 datasets spanning the 11 OpenReview primary-area buckets of the NeurIPS 2025 D\&B track, 97.4\% on 55 Open Targets datasets, 97.8\% on independent FHIR benchmarks, and 100\% strict tag-ID agreement against the DICOM PS3.6 dictionary across the \texttt{dcm\_validate} corpus, and bakes 51 OpenNeuro BIDS datasets without failure. The technical contribution underneath these numbers is the recovery layer: the format-specific knowledge required to assemble WFDB records, dispatch FHIR serializations, expand nested schemas, and parse imaging headers without materializing payloads. Treating that knowledge as a first-class extension surface, rather than work to hand off to upload-based platforms or to LLM agents alone, is what lets the modular handler architecture grow incrementally while the separation of metadata generation from data hosting preserves institutional compliance. As structured metadata becomes a prerequisite for dataset publication and review, tools that operate within governance boundaries will be essential to ensure that governed data can participate in standardized ML workflows on equal footing with platform-hosted data.

\textbf{Code availability.}
The source code, test suite (with bundled fixture datasets), and held-out evaluation scripts are available at \url{\ghlink}. Per-dataset access pointers, hosting platforms, and licenses for the evaluation corpora are listed in Appendix~\ref{app:data_sources}.


\clearpage

\bibliographystyle{plainnat}
\bibliography{references}

\appendix

\section{Design Goals and Generation Pipeline}
\label{app:design_goals}

Croissant Baker's design rests on five objectives: (i)~local-first execution without data upload or network connectivity; (ii)~minimal user burden via automatic structure and type inference; (iii)~standards compliance validated through the \texttt{mlcroissant} library~\citep{mlcroissant}; (iv)~extensibility through a typed handler protocol; and (v)~deterministic, auditable output via a clean separation of file-derived structural metadata from CLI-supplied semantic metadata. The structural layer covers file locations, content sizes, checksums, RecordSet schemas, and field data types; the semantic layer covers description, citation, license, creator, and publisher. Every value in the generated Croissant document is therefore traceable either to source-file bytes or to an explicit CLI input. The same separation makes Croissant Baker a composable building block for larger metadata workflows: an agent or downstream tool can propose semantic values without changing the extraction path that produces the structural layer~\citep{mlcommons2025eclair,anthropic2024mcp}. Algorithm~\ref{alg:pipeline} summarizes the four-stage pipeline.

\begin{algorithm}[ht]
\caption{Deterministic metadata generation pipeline. Structural assembly is
  kept distinct from semantic merge so that every field in the output is
  traceable to either source-file bytes or an explicit user input.}
\label{alg:pipeline}
\begin{algorithmic}[1]
\Require dataset directory $\mathcal{D}$, CLI-supplied semantic metadata $u$
\Ensure validated Croissant document
\State $\mathit{files} \gets \textsc{DiscoverFiles}(\mathcal{D})$
\State compute $\textsc{SHA256}$ and byte size for each $f \in \mathit{files}$
\For{$f \in \mathit{files}$}
    \State $h \gets \textsc{HandlerRegistry.dispatch}(f)$ \Comment{extension + content sniffing}
    \State $\mathit{record}[f] \gets h.\textsc{extract}(f)$ \Comment{schema, types, sub-file groupings}
\EndFor
\State $\mathit{structural} \gets \textsc{Assemble}(\mathit{record})$ \Comment{FileObjects, RecordSets, Fields}
\State $\mathit{croissant} \gets \textsc{Merge}(\mathit{structural}, u)$ \Comment{attach semantic CLI inputs}
\State \textsc{Validate}($\mathit{croissant}$)
\State \textsc{Serialize}($\mathit{croissant}$)
\end{algorithmic}
\end{algorithm}

\section{Representative CLI Invocation}
\label{app:cli_example}

Listing~\ref{lst:cli} shows an end-to-end invocation against the MIMIC-IV demo dataset. Structural metadata (file paths, checksums, types, RecordSet schemas) is inferred from \texttt{--input}; the remaining flags are semantic and Responsible AI (RAI) fields. The JSON-LD output corresponding to this invocation is shown in Appendix~\ref{app:jsonld}.

\begin{figure}[ht]
  \centering
  \begin{lstlisting}[basicstyle=\ttfamily\small,frame=single,framerule=0.4pt,
    breaklines=true,columns=flexible,keepspaces=true,
    showstringspaces=false,xleftmargin=0pt,xrightmargin=0pt]
$ pip install croissant-baker
$ croissant-baker \
    --input data/mimic-iv-demo \
    --output mimic-iv-demo.json \
    --name "MIMIC-IV Demo Dataset" \
    --description "Demo subset of MIMIC-IV." \
    --license "PhysioNet Restricted Health Data License 1.5.0" \
    --citation "Johnson et al., 2023" \
    --creator "Alistair Johnson" \
    --creator "Lucas Bulgarelli" \
    --creator "Tom Pollard" \
    --creator "Steven Horng" \
    --creator "Leo Anthony Celi" \
    --creator "Roger Mark" \
    --dataset-version "2.2" \
    --date-published "2023-01-06" \
    --url "https://physionet.org/content/mimic-iv-demo/2.2/" \
    --rai-data-use-cases "Clinical research and ML model development" \
    --rai-data-limitations "Demo subset; not for clinical decisions" \
    --rai-personal-sensitive-information "De-identified per HIPAA Safe Harbor"
  \end{lstlisting}
  \caption{\textbf{Installation and a representative invocation.}
    Structural metadata (file paths, checksums, types, RecordSet schemas)
    are inferred from \texttt{--input}; semantic fields and Responsible
    AI (RAI) attributes are passed through CLI flags. The corresponding
    JSON-LD output is in Appendix~\ref{app:jsonld}.}
  \label{lst:cli}
\end{figure}

\section{Format Handler Details}
\label{app:format_handlers}

This appendix expands on the built-in handlers summarized in
Section~\ref{sec:format_depth}.

\textbf{CSV/TSV Handler.} Supports plain CSV and TSV files and their compressed variants (\texttt{.csv.gz}, \texttt{.csv.bz2}, \texttt{.csv.xz}, \texttt{.tsv.gz}, \texttt{.tsv.bz2}, \texttt{.tsv.xz}). Column types are inferred via Apache Arrow type detection, supplemented by pattern recognition for temporal fields; Arrow types are mapped to Croissant equivalents. OMOP Common Data Model datasets are processed through this handler, as OMOP tables are distributed as compressed CSVs.

\textbf{Parquet Handler.} Reads Apache Arrow schema metadata without loading full datasets into memory, enabling efficient handling of large columnar files. MEDS-format datasets~\citep{mcdermott2025meds} are processed through this handler, as MEDS uses Parquet as its on-disk encoding.

\textbf{FHIR Handler.} Supports two FHIR~\citep{hl7fhir} serialization paths found in clinical research datasets: (i) NDJSON bulk export (\texttt{.ndjson}, \texttt{.ndjson.gz}), where each line is one resource of the same \texttt{resourceType} (FHIR Bulk Data specification); and (ii) JSON documents (\texttt{.json}, \texttt{.json.gz}) that either contain a Bundle resource whose \texttt{entry[]} array may mix resource types, or a single FHIR resource. The handler uses content sniffing to verify the presence of a \texttt{resourceType} key before accepting a file, so non-FHIR JSON files are not incorrectly identified.

\textbf{JSON/JSONL Handler.} Supports generic JSON files not identified as FHIR: JSON arrays, single objects, and JSON Lines (\texttt{.jsonl}, \texttt{.ndjson} with non-FHIR content), including gzip-compressed variants. The handler infers schema from inspected records.

\textbf{WFDB Handler.} Supports physiological waveform datasets in WFDB format~\citep{xie2023wfdb}. A WFDB record couples a header file (\texttt{.hea}), which encodes signal metadata such as channel names, sampling frequency, and record length, with one or more data files (\texttt{.dat}) and optional annotation files (\texttt{.atr}). This file-level coupling is domain-specific structure that generic ML metadata tools do not encode, leading to incomplete or missing RecordSets for waveform datasets (see Table~\ref{tab:hf_comparison}). The handler reads \texttt{.hea} headers via the WFDB Python library to parse sampling frequency, lead names, and record structure, then constructs FileObject entries for each component file alongside a RecordSet describing the logical signal structure.

\textbf{Image Handler.} Supports JPEG, PNG, TIFF, GIF, BMP, and WebP formats. For each image, the handler extracts dimensions, band count, and format, computes SHA-256 checksums, and returns image properties for Croissant generation. Standard RGB and grayscale images are processed using Pillow, while multi-band scientific TIFFs (e.g., 12-band Sentinel-2) are handled via \texttt{tifffile}, enabling unified metadata extraction for multimodal datasets combining images with tabular data.

\textbf{DICOM Handler.} Supports \texttt{.dcm} and \texttt{.dicom} files that contain the DICOM preamble and \texttt{DICM} magic bytes. The handler reads files using \texttt{pydicom} in header-only mode (\texttt{stop\_before\_pixels}) to avoid loading pixel arrays, while extracting modality, image geometry, frame count, pixel encoding, and core study/series identifiers. This matches the local-first design goal: structural metadata are inferred without moving data or materializing large imaging payloads.

\textbf{NIfTI Handler.} Supports \texttt{.nii} and \texttt{.nii.gz} volumes using \texttt{nibabel} in header-only mode. The handler reads file headers to extract spatial dimensions, voxel spacing, data type, NIfTI version, and repetition time for 4D acquisitions. This design enables Croissant metadata generation for neuroimaging-style datasets without loading dense voxel arrays into memory.

\section{Held-Out Evaluation Details}
\label{app:heldout_eval_details}

This appendix expands on the four established held-out splits summarized in
Table~\ref{tab:heldout_summary} (the cross-domain NeurIPS 2025 split is described
in Section~\ref{sec:neurips_2025_cross_domain}).

\subsection{External Validation Against Producer-Authored Metadata}
\label{sec:external_validation}

To evaluate type inference accuracy against human expert annotation, we compare Croissant Baker outputs with producer-authored Croissant metadata released by Open Targets~\citep{opentargets_croissant}. Open Targets provides a curated \texttt{croissant.json} for each dataset, serving as ground-truth metadata for a large, multi-dataset genomics resource.

We download all Parquet partitions for each of the 55 datasets (approximately 20--30\,GB in total, depending on the current Open Targets release) and run Croissant Baker on the full local copy. Croissant Baker recovers all 55 RecordSet names and all 819 field names, with 798/819 semantic types matching the producer-authored metadata (97.4\%). Appendix Table~\ref{tab:opentargets} presents the detailed counts.

The remaining 21 disagreements (2.6\%) arise when the human-authored metadata assigns \texttt{sc:Float} to integer-valued columns, whereas Croissant Baker infers the more precise integer type directly from the Parquet schema. We observe one structural difference: the producer-authored metadata uses \texttt{cr:FileSet} with glob patterns to represent partitioned directories, while Croissant Baker produces individual \texttt{cr:FileObject} entries per file. This difference does not affect schema-level accuracy. Detailed counts appear in Table~\ref{tab:opentargets}.

\begin{table}[ht]
\caption{Open Targets validation against producer-authored metadata (55 Parquet datasets).}
\label{tab:opentargets}
\centering
\small
\begin{tabular}{lrr}
\toprule
\textbf{Metric} & \textbf{Count} & \textbf{Agreement} \\
\midrule
RecordSet name recovery & 55 / 55  & 100\% \\
Field name recovery     & 819 / 819 & 100\% \\
Semantic type agreement & 798 / 819 & 97.4\% \\
\bottomrule
\end{tabular}
\end{table}

\subsection{FHIR Out-of-Distribution Evaluation}
\label{sec:fhir_ood}

To evaluate FHIR handler generalization beyond the MIMIC-IV FHIR data used during development, we apply Croissant Baker to two independent FHIR sources from the SMART Health IT project~\citep{smart_custom_sample} (Boston Children's Hospital / Harvard), covering both FHIR wire formats supported by the handler.

\textbf{NDJSON bulk export.} We process the 10-patient synthetic FHIR NDJSON bulk export sample~\citep{smart_bulk_fhir} (18 NDJSON files). Croissant Baker generates 18 RecordSets with 186 top-level fields (406 leaf fields after expanding nested structures such as \texttt{address} and \texttt{valueQuantity}) in 2.6\,s. All outputs pass \texttt{mlcroissant} validation, including correct multi-chunk merging of Observation data across two NDJSON files.

To assess type inference accuracy, we construct a standards-grounded reference by resolving each observed leaf field against the applicable US Core STU7 profile~\citep{hl7uscore7} (using \texttt{meta.profile} declarations) and falling back to base HL7 FHIR~R4 StructureDefinitions~\citep{hl7fhir} for nested datatypes and uncovered resource types. All 406 leaf paths resolve to the specification (0 unresolved paths). Croissant Baker achieves 397/406 strict type agreement (97.8\%) across all 18 resource types. The 9 disagreements fall into two categories. First, eight cases arise where the FHIR specification represents URI fields using the FHIRPath \texttt{System.String} primitive (ground truth: \texttt{sc:Text}), while Croissant Baker infers the more specific \texttt{sc:URL} from observed URL-shaped content. Second, one case occurs where a decimal field contains only integer values in the 10-patient sample, leading Croissant Baker to infer \texttt{cr:Int64} rather than the specification's \texttt{cr:Float64}.

\textbf{JSON Bundle format.} We process 5 FHIR transaction Bundle JSON files~\citep{smart_custom_sample}, exercising the Bundle ingestion path. Croissant Baker groups resources across the Bundle files by type, producing 11 RecordSets (94 total fields) backed by a single FileSet. All outputs pass \texttt{mlcroissant} validation.

\subsection{Paired DICOM and NIfTI Out-of-Distribution Evaluation}
\label{sec:imaging_ood}

To evaluate DICOM and NIfTI handler generalization beyond development fixtures, we apply Croissant Baker to the \texttt{dcm\_validate} corpus~\citep{rorden2025dcmvalidate}, a published reference dataset that pairs vendor DICOM acquisitions with their NIfTI conversions and BIDS JSON sidecars. We process six modules covering five MRI vendors and one cross-vendor enhanced DICOM module. Across these modules, Croissant Baker bakes 6{,}678 DICOM files and 75 NIfTI volumes, while lifting 3{,}932 BIDS sidecar fields through the JSON handler within the same bake. All outputs pass Croissant~1.1 validation via \texttt{mlcroissant}.

To assess DICOM type inference accuracy, we resolve each DICOM tag identifier emitted in field descriptions against the DICOM PS3.6 data dictionary~\citep{dicom_ps36} via \texttt{pydicom.datadict}. All 48 emitted tag identifiers across the six modules resolve to valid PS3.6 keywords (100\% strict tag-ID agreement). Appendix~\ref{app:imaging_listings} reports per-module file counts.

\subsection{NIfTI Handler Validation at Scale via OpenNeuro}
\label{sec:openneuro_ood}

To evaluate the NIfTI handler generalization across heterogeneous independent datasets, we apply Croissant Baker to 51 publicly available BIDS datasets from OpenNeuro~\citep{openneuro2022}. The set includes 50 of the smallest publicly accessible datasets in the bucket, plus a reference fMRI dataset (\texttt{ds000003}, Rhyme Judgment) to ensure functional-acquisition coverage. None of these datasets are used during development. The combined corpus spans approximately 1.66\,GB across T1-weighted anatomy, fMRI, PET, and signal-only or behavioral releases.

Croissant Baker bakes all 51 datasets without failure. Among these, 36 datasets contain NIfTI volumes (268 volumes total) and 15 contain only sidecar JSON or tabular files. The JSON handler lifts 7{,}978 RecordSets from BIDS sidecars and standalone metadata; the TSV handler emits 13{,}966 tabular RecordSets across BIDS \texttt{events.tsv}, \texttt{participants.tsv}, and related files. All outputs pass Croissant~1.1 validation via mlcroissant. The NIfTI, JSON, and TSV handlers operate concurrently on each dataset without dataset-specific configuration. We list the 51 dataset identifiers in Appendix~\ref{app:imaging_listings}.

\section{Evaluation Splits}
\label{app:eval_splits_table}

Table~\ref{tab:eval_splits} enumerates each split's corpus, whether it appears during development, the ground-truth source, and the claim it supports. The two local splits at the top establish coverage and scale; the five held-out splits below (NeurIPS 2025 cross-domain, Open Targets, FHIR, paired DICOM/NIfTI, and OpenNeuro) are external to development and are detailed in Section~\ref{sec:neurips_2025_cross_domain} and Appendix~\ref{app:heldout_eval_details}.

\begin{table}[ht]
\caption{Evaluation splits and the claims they support.}
\label{tab:eval_splits}
\centering
\scriptsize
\setlength{\tabcolsep}{3pt}
\begin{tabular}{@{}>{\raggedright\arraybackslash}p{0.16\linewidth}
                >{\raggedright\arraybackslash}p{0.21\linewidth}
                >{\raggedright\arraybackslash}p{0.11\linewidth}
                >{\raggedright\arraybackslash}p{0.13\linewidth}
                >{\raggedright\arraybackslash}p{0.22\linewidth}@{}}
\toprule
\textbf{Split} & \textbf{Datasets} & \textbf{Used in development} & \textbf{Ground truth} & \textbf{Claim supported} \\
\midrule
Development coverage & 7 local datasets spanning CSV/TSV, Parquet, WFDB, OMOP-through-CSV, and multimodal image + tabular data & Yes & No & End-to-end validity on representative real-world formats \\
Scalability validation & MIMIC-IV full and MIMIC-IV MEDS full & No & No & Runtime and repository scale on full institutional data \\
NeurIPS 2025 cross-domain & 25 datasets across 11 OpenReview primary-area buckets (Fig.~\ref{fig:dandb_landscape}), 3 seeds & No & Croissant snapshots submitted with each paper at review time & Cross-domain coverage of the NeurIPS 2025 D\&B track \\
External validation & Open Targets (55 Parquet datasets) & No & Producer-authored Croissant metadata & RecordSet recovery, field-name recovery, and type agreement \\
Independent FHIR evaluation & SMART bulk NDJSON and SMART JSON Bundle files & No & Expected structural outcomes & Generalization to an independent source and a second ingestion path \\
Paired DICOM and NIfTI & \texttt{dcm\_validate} corpus, 6 vendor modules & No & DICOM PS3.6 data dictionary (\texttt{pydicom.datadict}) & Cross-vendor DICOM and NIfTI handler generalization \\
NIfTI handler at scale & OpenNeuro, 51 BIDS datasets & No & mlcroissant 1.1 validity and BIDS structural conformance & NIfTI handler generalization across heterogeneous independent datasets \\
\bottomrule
\end{tabular}
\end{table}

\section{Local Evaluation Dataset Inventory}
\label{app:local_datasets}

Table~\ref{tab:results} reports per-dataset file counts, FileObject and RecordSet totals, repository size, and end-to-end generation time for the nine local datasets covered by the development and scalability splits. The seven smaller datasets exercise the CSV/TSV, Parquet, WFDB, and image handlers; MIMIC-IV full and MIMIC-IV MEDS full are the two institutional-scale repositories reserved for runtime validation.

\begin{table}[ht]
\caption{Croissant Baker outputs for the nine development and scalability datasets. Reported time includes file discovery, SHA-256 checksum computation, type inference, and JSON-LD serialization; all outputs pass \texttt{mlcroissant} validation.}
\label{tab:results}
\centering
\small
\begin{tabular}{lrrrrr}
\toprule
\textbf{Dataset} & \textbf{Files} & \textbf{FileObjects} & \textbf{RecordSets} & \textbf{Size} & \textbf{Time} \\
\midrule
MIMIC-IV Demo v2.2          &  66 &  66 &  59 & 287\,MB  & 2.73\,s \\
eICU-CRD Demo v2.0.1        &  31 &  31 &  31 &  52\,MB  & 2.09\,s \\
MIT-BIH Arrhythmia v1.0.0   & 213 & 213 &  71 & 107\,MB  & 1.60\,s \\
MIMIC-IV OMOP v0.9          &  32 &  32 &  32 &  66\,MB  & 1.80\,s \\
MIMIC-IV MEDS Demo v0.0.1   &   5 &   5 &   5 & 5.8\,MB  & 0.88\,s \\
Glaucoma Fundus (HYGD)      &  13 &  13 &   2 & 1.4\,MB  & 0.74\,s \\
Satellite Public Health     &  11 &  11 &   2 &  33\,MB  & 3.30\,s \\
MIMIC-IV Full (CSV)         &  32$^{*}$ &  31 &  31 & 9.92\,GB & 13.3\,s \\
MIMIC-IV MEDS Full (Parquet)& 374$^{*}$ & 366 & 366 & 3.67\,GB & 32.2\,s \\
\midrule
\textbf{Total}              & 777 & 768 & 599 & --- & --- \\
\bottomrule
\end{tabular}
\begin{minipage}{\linewidth}
\vspace{4pt}
\footnotesize $^{*}$\,Files counts all files found by recursive traversal; FileObjects counts only files matched by a format handler. The difference corresponds to non-data files present in the directory (README, checksums, or Parquet partition control files) that produce no Croissant artifact.
\end{minipage}
\end{table}

\section{Proposed Agent-Assisted Semantic Enrichment Extension}
\label{app:agent_extension}

\begin{figure}[ht]
  \centering
  \includegraphics[width=\linewidth]{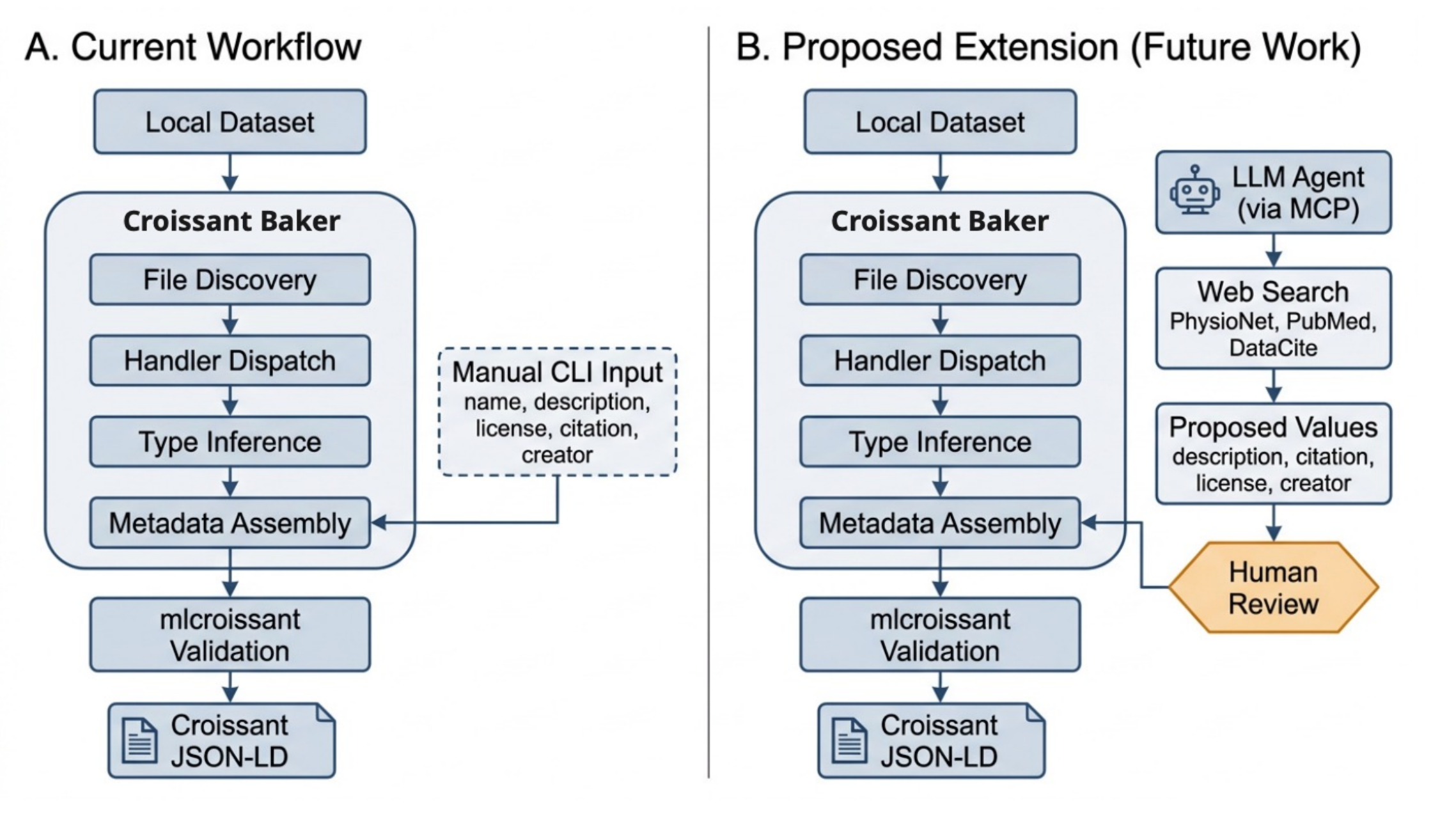}
  \caption{\textbf{Proposed agent-assisted metadata enrichment extension.} (A) Current workflow: a deterministic pipeline performs file discovery, handler dispatch, type inference, and metadata assembly, while semantic fields (description, citation, license, creator) are provided manually via CLI. (B) Proposed future extension: an LLM agent queries publicly available documentation to retrieve and propose values for non-inferable fields, subject to mandatory human review. The deterministic structural core remains unchanged; the underlying dataset never leaves the local environment.}
  \label{fig:agent_extension}
\end{figure}

Figure~\ref{fig:agent_extension} sketches the agent-assisted extension discussed at the end of the Discussion section: an LLM agent operating through the Model Context Protocol proposes values for the semantic fields that Croissant Baker currently requires as CLI input (description, citation, license, creator), subject to mandatory human review. The deterministic structural core (file discovery, handler dispatch, type inference, metadata assembly) is unchanged, and the dataset itself never leaves the local environment.

\section{Per-Dataset Comparison with Hugging Face and Kaggle}
\label{app:hf_kaggle}

We upload seven open-access evaluation datasets to Hugging Face and Kaggle and compare auto-generated Croissant metadata with Croissant Baker output. The generated JSON-LD files used in this comparison are provided as supplementary material.

\textbf{Summary of comparison themes.}
(1)~\emph{Per-file integrity:} Croissant Baker provides SHA-256 checksums and exact byte sizes for every file; HF provides a placeholder link; Kaggle provides one MD5 checksum for the entire archive only. (2)~\emph{Precise data types:} PyArrow-inferred types (\texttt{cr:Int64}, \texttt{cr:Float64}, \texttt{sc:DateTime}) match or exceed HF precision and surpass Kaggle (coarse \texttt{sc:Integer}/\texttt{sc:Float}; integer IDs are misclassified as \texttt{sc:Text}). (3)~\emph{Multi-band and multimodal:} For 12-band Sentinel-2 TIFFs and glaucoma fundus datasets (tabular + images), HF produces empty or merged metadata; Kaggle detects files but no structural RecordSets for images; Croissant Baker instead generates per-file FileObjects, separate RecordSets per modality, and image summary RecordSets. (4)~\emph{Spec compliance and provenance:} Croissant Baker produces \texttt{mlcroissant}-valid output with license, citation, datePublished, and version; HF/Kaggle do not auto-generate these.

\begin{table}[ht]
\caption{Croissant Baker vs.\ Hugging Face (evaluation subsets).}
\label{tab:hf_comparison}
\centering
\footnotesize
\begin{tabular}{p{0.20\linewidth}p{0.70\linewidth}}
\toprule
\textbf{Dataset} & \textbf{Comparison} \\
\midrule
eICU demo &
  HF auto-generation does not produce RecordSets or fields; Croissant Baker
  produces 30 RecordSets and 390 fields with inferred types (Int64, Float64,
  Text, Boolean, Time), along with file sizes, SHA-256 checksums, row counts, and license/citation/date. \\
\addlinespace
MIMIC-IV demo &
  HF does not produce RecordSets/fields; Croissant Baker produced 59 RecordSets
  and 428 fields with precise types (Int64, Float64, DateTime), per-file
  SHA-256, row counts, and provenance. \\
\addlinespace
MIMIC-IV MEDS demo &
  HF represents 5 Parquet files as one ``train'' split with 25 fields;
  Croissant Baker describes 5 separate RecordSets with 83 total fields, plus
  SHA-256 and contentSize per file and row counts. \\
\addlinespace
MIMIC-IV OMOP demo &
  HF does not produce RecordSets/fields; Croissant Baker produced 32 RecordSets
  and 367 fields (OMOP tables) with types, SHA-256, row counts, and
  CC~BY~4.0/citation/datePublished. \\
\addlinespace
MIT-BIH WFDB demo &
  HF produces a schema.org Dataset without Croissant RecordSets; Croissant
  Baker produces 213 FileObjects, 71 RecordSets, and 142 fields, reading
  \texttt{.hea} headers for sampling rate and signal names, with SHA-256,
  contentSize, and license/citation/date. \\
\addlinespace
Satellite Public Health &
  HF does not produce RecordSets or distributions for multi-band TIFFs or
  CSV; Croissant Baker produces 11 FileObjects (1 CSV + 10 TIFFs) with
  SHA-256 and sizes, a tabular RecordSet for all 1017 columns, and an image
  summary RecordSet for the 12 Sentinel-2 bands. \\
\addlinespace
Glaucoma Fundus &
  HF converts images to Parquet (one ``train'' split); Croissant Baker lists
  12 JPGs and Labels.csv as separate FileObjects with SHA-256 and sizes, a
  tabular RecordSet for Labels.csv (\texttt{cr:Int64}, \texttt{cr:Float64}),
  and an image summary RecordSet. \\
\bottomrule
\end{tabular}
\end{table}

\begin{table}[ht]
\caption{Croissant Baker vs.\ Kaggle (evaluation subsets).}
\label{tab:kaggle_comparison}
\centering
\footnotesize
\begin{tabular}{p{0.20\linewidth}p{0.70\linewidth}}
\toprule
\textbf{Dataset} & \textbf{Comparison} \\
\midrule
eICU demo &
  Both cover 31 RecordSets and 391 fields. Kaggle assigns 4 timestamp-related columns as \texttt{sc:Date} (eICU stores minutes-from-admission); Croissant Baker correctly infers \texttt{cr:Int64}. Kaggle uses coarse \texttt{sc:Integer}/\texttt{sc:Float}; Croissant Baker uses \texttt{cr:Int64}/\texttt{cr:Float64} and detects \texttt{sc:Time} for time-of-day. Kaggle provides one MD5 checksum on the archive; Croissant Baker provides SHA-256 per file with per-file contentSize and row counts. \\
\addlinespace
MIMIC-IV demo &
  Kaggle assigns many ID columns (\texttt{subject\_id}, \texttt{hadm\_id}) as \texttt{sc:Text}; Croissant Baker infers \texttt{cr:Int64}. Kaggle types all date/time columns as \texttt{sc:Date}; Croissant Baker distinguishes \texttt{sc:DateTime} from \texttt{sc:Date}. Kaggle provides one MD5 checksum on zip; Croissant Baker provides SHA-256 per file with sizes and row counts. \\
\addlinespace
MIMIC-IV MEDS demo &
  Kaggle does not produce RecordSets/fields for Parquet files; Croissant Baker produces 5 RecordSets with 83 fields, 5 FileObjects with SHA-256 and contentSize, and row counts. \\
\addlinespace
MIMIC-IV OMOP demo &
  Kaggle reports 32 RecordSets (367 fields) including 10 empty tables; Croissant Baker reports 32 RecordSets (367 fields). Kaggle detects \texttt{sc:URL} for URL columns in vocabulary tables (future work for Croissant Baker). Kaggle assigns all date/time as \texttt{sc:Date}; Croissant Baker distinguishes \texttt{sc:DateTime}. \\
\addlinespace
MIT-BIH WFDB demo &
  Kaggle does not produce RecordSets/fields for WFDB signal files; Croissant Baker produces 71 RecordSets, 142 fields, and 213 FileObjects with SHA-256, contentSize, and \texttt{.hea}-derived metadata (360\,Hz, MLII/V5, \texttt{sc:Float}). \\
\addlinespace
Satellite Public Health &
  Kaggle provides archive with MD5 and \texttt{cr:FileSet} for \texttt{.tiff}/\texttt{.csv} but no RecordSets or field definitions. Croissant Baker produces column-level metadata for all 1017 CSV columns and an image summary RecordSet for the TIFFs (12-band). \\
\addlinespace
Glaucoma Fundus &
  Kaggle provides archive with MD5; Labels.csv has coarse types and there is a \texttt{cr:FileSet} for \texttt{*.jpg} but no image RecordSet. Croissant Baker lists all 13 files with SHA-256, Labels with \texttt{cr:Int64}/ \texttt{cr:Float64}, and an \texttt{sc:ImageObject} summary RecordSet. \\
\bottomrule
\end{tabular}
\end{table}

\section{Downstream Utility Assessment (Detailed)}
\label{app:downstream}

\begin{enumerate}
  \item \textbf{Programmatic loading without dataset-specific code.}
        Using the \texttt{mlcroissant} Python API, we load generated Croissant files for each dataset and iterate over RecordSet objects. For tabular datasets, this yields typed dictionaries. For WFDB waveform data, iteration produces logical ECG records that correctly associate header and signal files, demonstrating schema-aware iteration with ingestion logic decoupled from directory conventions.

  \item \textbf{Automated integrity and packaging verification.}
        Controlled perturbations---removal of one referenced file, renaming of a waveform component, modification of a column name---are all detected by mlcroissant validation before downstream analysis. This supports pre-release packaging checks and continuous integration for institutional repositories.

  \item \textbf{Cross-site schema verification.}
        For OMOP and MEDS datasets, we programmatically extract schema definitions and compare table presence, column counts, and type mappings across datasets. This enables rapid verification that two institutions claiming OMOP compatibility expose equivalent schemas prior to federated modeling.

  \item \textbf{Metadata-only sharing for controlled-access datasets.}
        For MIMIC-IV, Croissant Baker generates metadata entirely locally. The JSON-LD files include dataset-level descriptors, file distributions with checksums, and RecordSet definitions, without exposing patient-level data---enabling public discoverability while preserving access restrictions.
\end{enumerate}

\section{Evaluation Subset Composition (Image Datasets)}
\label{app:subsets}

\textbf{Glaucoma Fundus (HYGD).}
Source: Hillel Yaffe Glaucoma Dataset (HYGD) v1.0.0~\citep{physionet_hygd}
(ODbL v1.0). Full dataset: 747 JPG fundus images from 304 subjects
(\textasciitilde126\,MB). Evaluation subset: 12 JPGs from 12 subjects
(8~GON+, 4~GON--), with Labels.csv filtered to those 12 rows.

\textbf{Satellite Public Health.}
Source: ``A Multi-Modal Satellite Imagery Dataset for Public Health
Analysis in Colombia'' v1.0.0~\citep{physionet_satellite} (CC0~1.0). Full
dataset: 65\,GB (81 municipalities, \textasciitilde12,636 Sentinel-2 TIFFs,
2016--2018). Evaluation subset: 10 TIFFs from 2 municipalities (5001
Medell\'{\i}n, 8001 Barranquilla), 5 images each over January--July~2016;
metadata.csv filtered to the 10 corresponding rows. Images are 12-band
Sentinel-2 TIFFs (\textasciitilde745$\times$747\,px, uint8); metadata.csv
has 1017 columns. Directory structure and column set are unchanged.

\section{Benchmark Machine Specifications}
\label{app:hardware}

All automated timing benchmarks reported in Table~\ref{tab:results}---for
both evaluation subsets and full-scale datasets---are executed locally on
a MacBook Pro with an Apple M1~Max processor (10~cores) and 32\,GB RAM
running macOS. No cloud compute is used.

\clearpage
\section{Example JSON-LD Metadata}
\label{app:jsonld}

The following snippet shows Croissant Baker output for the
\texttt{admissions} table from MIMIC-IV Demo, demonstrating dataset-level
provenance, a FileObject with SHA-256 checksum, and a RecordSet with
automatically inferred column types.

\begin{lstlisting}[language=json]
{
  "@context": {
    "@language": "en",
    "@vocab": "https://schema.org/",
    "cr": "http://mlcommons.org/croissant/",
    "rai": "http://mlcommons.org/croissant/RAI/",
    "sc": "https://schema.org/"
  },
  "@type": "sc:Dataset",
  "conformsTo": [
    "http://mlcommons.org/croissant/1.1",
    "http://mlcommons.org/croissant/RAI/1.0"
  ],
  "name": "MIMIC-IV Demo Dataset",
  "description": "Demo subset of MIMIC-IV.",
  "license": "PhysioNet Restricted Health Data License 1.5.0",
  "version": "2.2",
  "datePublished": "2023-01-06",
  "url": "https://physionet.org/content/mimic-iv-demo/2.2/",
  "creator": [
    { "@type": "sc:Person", "name": "Alistair Johnson" },
    { "@type": "sc:Person", "name": "Lucas Bulgarelli" },
    { "@type": "sc:Person", "name": "Tom Pollard" },
    { "@type": "sc:Person", "name": "Steven Horng" },
    { "@type": "sc:Person", "name": "Leo Anthony Celi" },
    { "@type": "sc:Person", "name": "Roger Mark" }
  ],
  "citeAs": "Johnson et al., 2023",
  "rai:dataUseCases": "Clinical research and ML model development",
  "rai:dataLimitations": "Demo subset; not for clinical decisions",
  "rai:personalSensitiveInformation": "De-identified per HIPAA Safe Harbor",
  "distribution": [{
    "@type": "cr:FileObject",
    "@id": "file_13",
    "name": "admissions.csv.gz",
    "contentSize": "11072",
    "contentUrl": "hosp/admissions.csv.gz",
    "encodingFormat": "application/gzip",
    "sha256": "910b9f160ffdf1e08ea673585393f347c773ccc87d66875c627584a903ae8493"
  }],
  "recordSet": [{
    "@type": "cr:RecordSet",
    "@id": "recordset_13",
    "name": "admissions",
    "field": [
      {
        "@type": "cr:Field",
        "@id": "file_13_subject_id",
        "name": "subject_id",
        "dataType": "cr:Int64",
        "source": {
          "fileObject": { "@id": "file_13" },
          "extract": { "column": "subject_id" }
        }
      },
      {
        "@type": "cr:Field",
        "@id": "file_13_admittime",
        "name": "admittime",
        "dataType": "sc:DateTime",
        "source": {
          "fileObject": { "@id": "file_13" },
          "extract": { "column": "admittime" }
        }
      }
    ]
  }]
}
\end{lstlisting}

\section{NeurIPS 2025 Datasets and Benchmarks Composition Data}
\label{app:dandb_composition}

The Sankey figure in Section~\ref{sec:neurips_2025_cross_domain} (Figure~\ref{fig:dandb_landscape})
visualizes the relationship between paper domain and dataset hosting
platform across the 497 papers accepted to the NeurIPS 2025 Datasets and
Benchmarks track. Domain assignments come from the OpenReview
\texttt{primary\_area} field; hosting platforms come from the
\texttt{dataset\_URL} field, normalized to the canonical platform names
listed below. Croissant availability is checked by attempting to fetch a
machine-readable Croissant document from each host.

Across all 497 accepted papers, 426 (85.7\%) include a retrievable Croissant
document and 71 (14.3\%) do not. Tables~\ref{tab:dandb_domain_totals}
and~\ref{tab:dandb_host_totals} report the domain-side and platform-side
totals; Table~\ref{tab:dandb_top_flows} reports the largest
domain~$\rightarrow$~platform flows.

\begin{table}[h]
\caption{Per-domain paper counts in the NeurIPS 2025 D\&B track
  (\(N=497\)).}
\label{tab:dandb_domain_totals}
\centering
\small
\begin{tabular}{lrr}
\toprule
\textbf{Domain (primary\_area)} & \textbf{Papers} & \textbf{With Croissant} \\
\midrule
Language / LLM                & 165 & 152 \\
Computer Vision               & 69  & 61  \\
Other domains                 & 65  & 53  \\
Evaluation                    & 36  & 27  \\
Health sciences               & 29  & 28  \\
Physics                       & 28  & 27  \\
Life sciences                 & 27  & 26  \\
Deep-learning scenarios       & 25  & 17  \\
Social / economic             & 24  & 16  \\
Reinforcement learning        & 16  & 6   \\
Speech / audio                & 13  & 13  \\
\midrule
\textbf{Total}                & \textbf{497} & \textbf{426 (85.7\%)} \\
\bottomrule
\end{tabular}
\end{table}

\begin{table}[h]
\caption{Per-platform paper counts in the NeurIPS 2025 D\&B track.}
\label{tab:dandb_host_totals}
\centering
\small
\begin{tabular}{lrr}
\toprule
\textbf{Hosting platform} & \textbf{Papers} & \textbf{With Croissant} \\
\midrule
Hugging Face          & 321 & 319 \\
No public dataset URL & 72  & 3   \\
Kaggle                & 36  & 36  \\
DOI / other registry  & 34  & 34  \\
GitHub                & 19  & 19  \\
Dataverse             & 11  & 11  \\
Zenodo                & 4   & 4   \\
\midrule
\textbf{Total}        & \textbf{497} & \textbf{426 (85.7\%)} \\
\bottomrule
\end{tabular}
\end{table}

\begin{table}[h]
\caption{Largest domain~$\rightarrow$~platform flows in the NeurIPS
  2025 D\&B track. Top 12 of 58 unique flows.}
\label{tab:dandb_top_flows}
\centering
\small
\begin{tabular}{llrr}
\toprule
\textbf{Domain} & \textbf{Platform} & \textbf{Papers} & \textbf{With Croissant} \\
\midrule
Language / LLM            & Hugging Face        & 135 & 134 \\
Computer Vision           & Hugging Face        & 43  & 43  \\
Other domains             & Hugging Face        & 34  & 34  \\
Evaluation                & Hugging Face        & 20  & 19  \\
Life sciences             & Hugging Face        & 19  & 19  \\
Health sciences           & Hugging Face        & 17  & 17  \\
Deep-learning scenarios   & Hugging Face        & 13  & 13  \\
Physics                   & Hugging Face        & 13  & 13  \\
Other domains             & No dataset URL      & 12  & 0   \\
Social / economic         & Hugging Face        & 12  & 12  \\
Language / LLM            & No dataset URL      & 12  & 0   \\
Speech / audio            & Hugging Face        & 11  & 11  \\
\bottomrule
\end{tabular}
\end{table}

\section{NeurIPS 2025 Cross-Domain Draw Protocol}
\label{app:neurips_2025_protocol}

This appendix records the design decisions behind the cross-domain
split in Section~\ref{sec:neurips_2025_cross_domain}, drawn from the
OpenReview snapshot of accepted NeurIPS 2025 D\&B papers summarized in
Figure~\ref{fig:dandb_landscape}.

A single random pick per \texttt{primary\_area} bucket gives only 11
datasets. To exercise more handlers per bucket and to make per-bucket
behavior less dependent on a particular shuffle, we draw three
independent seeds, producing 33 picks that resolve to 25 unique
datasets after removing cross-seed repeats. The three seed values are
fixed in the evaluation manifest distributed with the source code.

A pilot draw across all hosts (Hugging Face, Kaggle, GitHub, Zenodo,
Dataverse, DOI registries, PhysioNet) surfaced two reproducibility
frictions unrelated to Croissant Baker itself: one Kaggle pick had been
removed by its uploader before evaluation, and one PhysioNet pick
required credentialed-access onboarding (PhysioNet training and a
signed data use agreement) and overlapped the biomedical formats
already exercised in the FHIR and imaging splits. Restricting
eligibility to publicly retrievable Hugging Face datasets removes both
frictions, keeps the protocol re-runnable from a single
\texttt{huggingface-cli} login, and does not narrow modality coverage
relative to the pilot since Hugging Face hosts about two thirds of the
public datasets in the track.

The repository-size band of $[1\,\text{MB}, 3\,\text{GB}]$ is set for
breadth across the 33 picks within a practical compute and download
budget. Broader-scale repositories are covered separately in the
scalability split.

The candidate pool, the primary-area-to-bucket mapping, the three seed
values, and the deterministic \texttt{random.Random(seed)} shuffle
that resolves to the 33 picks are all included in the source release.
The per-pick test harness runs Croissant Baker on each pick and
validates each output with \texttt{mlcroissant} before computing
field-name and type agreement against the producer Croissants.

\section{External Imaging Evaluation Dataset Listings}
\label{app:imaging_listings}

This appendix lists the per-dataset identifiers used in the imaging
out-of-distribution evaluations (\S\ref{sec:imaging_ood} and
\S\ref{sec:openneuro_ood}) so that readers can reproduce the bakes
exactly.

\textbf{\texttt{dcm\_validate} modules.} Six modules from the Rorden et
al.\ corpus~\citep{rorden2025dcmvalidate}, each cloned from its
\texttt{neurolabusc/dcm\_qa\_*} GitHub mirror:

\begin{table}[h]
\caption{\texttt{dcm\_validate} modules processed in
  \S\ref{sec:imaging_ood}.}
\label{tab:dcm_validate_modules}
\centering
\small
\begin{tabular}{llrr}
\toprule
\textbf{Module} & \textbf{Vendor / variant} & \textbf{DICOM files} & \textbf{NIfTI volumes} \\
\midrule
\texttt{dcm\_qa\_ge}     & GE MRI                              & 4{,}125 & 29 \\
\texttt{dcm\_qa\_nih}    & NIH multi-vendor (GE + Siemens)     & 608     & 8  \\
\texttt{dcm\_qa\_polar}  & GE EPI phase-encoding variant       & 960     & 6  \\
\texttt{dcm\_qa\_stc}    & GE slice-timing reference           & 589     & 18 \\
\texttt{dcm\_qa\_uih}    & United Imaging Healthcare MRI       & 388     & 6  \\
\texttt{dcm\_qa\_enh}    & Enhanced DICOM cross-vendor         & 8       & 8  \\
\midrule
\textbf{Total}           &                                     & \textbf{6{,}678} & \textbf{75} \\
\bottomrule
\end{tabular}
\end{table}

\textbf{OpenNeuro dataset identifiers.} The 51 BIDS datasets processed
in \S\ref{sec:openneuro_ood}, in OpenNeuro accession order:

\begin{quote}
\small\ttfamily
ds000003, ds000204, ds001037, ds001450, ds001600,\\
ds001728, ds001780, ds002014, ds002040, ds002293,\\
ds002328, ds002374, ds002614, ds002733, ds002743,\\
ds002868, ds002873, ds002912, ds002936, ds002939,\\
ds002946, ds002982, ds002990, ds003098, ds003325,\\
ds003538, ds003805, ds003810, ds004129, ds004130,\\
ds004131, ds004215, ds004339, ds004552, ds004776,\\
ds004850, ds004853, ds004854, ds004855, ds004872,\\
ds005274, ds005412, ds005872, ds005929, ds005964,\\
ds006138, ds006392, ds006462, ds006486, ds007338,\\
ds007406
\end{quote}

Each OpenNeuro dataset is publicly accessible at
\url{https://openneuro.org/datasets/<id>}; bulk download proceeds
through the public S3 bucket \texttt{s3://openneuro.org/<id>/} via
path-style HTTPS URLs.

\section{Data Sources and Access}
\label{app:data_sources}

Croissant Baker is evaluated on publicly distributed third-party datasets; no new datasets are introduced by this work. Per-source access pointers and licensing terms are summarised here.

PhysioNet datasets (MIMIC-IV, Hillel Yaffe Glaucoma, and Multimodal Satellite Data) are available through PhysioNet (\href{https://physionet.org}{\texttt{physionet.org}}) under their respective data use agreements~\citep{PhysioNet2026}.
Open Targets Parquet datasets are available from the Open Targets Platform (\href{https://platform.opentargets.org}{\texttt{platform.opentargets.org}})~\citep{opentargets_platform}.
FHIR evaluation samples are available from SMART Health IT~\citep{smart_bulk_fhir,smart_custom_sample}.
The \texttt{dcm\_validate} DICOM/NIfTI corpus is available via Zenodo~\citep{rorden2025dcmvalidate}.
OpenNeuro BIDS datasets are available from \href{https://openneuro.org}{\texttt{openneuro.org}}~\citep{openneuro2022}.
NeurIPS 2025 cross-domain datasets are publicly available on Hugging Face; the draw protocol and candidate pool are released with the source code.


\end{document}